\documentstyle[11pt,a4,fullname]{article}
\title{Foreground and Background Lexicons and Word Sense
Disambiguation for Information Extraction} 

\author{Adam Kilgarriff\thanks{The work was undertaken under EPSRC grant GR/K/18931}
\\
Information Technology Research Institute\\
University of Brighton\\
Brighton BN2 4GJ\\
\tt email: Adam.Kilgarriff@itri.bton.ac.uk}

\date{}
\begin{document}
\maketitle
\section{Introduction}

In recent years, lexicon acquisition from machine-readable
dictionaries and corpora has been a dynamic field of research.
However it has not always been evident how lexical information so
acquired can be used, or how it relates to more structured meaning
representations.  In this paper I look at this issue in relation to one
particular NLP task, Information Extraction (hereafter IE), and
one subtask for which both lexical and general knowledge are required,
Word Sense Disambiguation (WSD).

The argument is as follows.  For an IE task, the output formalism,
that is, the database fields or templates which the system is to fill,
specifies the object-types and relations that the system is to find
out about; the `ontology'.  An IE task operates in a specific domain.
The task requires the key terms of that domain, the `foreground
lexicon', to be tightly bound to the ontology.  This is a task that
calls for human input.  For all other vocabulary, the `background
lexicon', a far shallower semantics will be sufficient.  This shallow
semantics can be obtained automatically from sources such as
machine-readable dictionaries and domain corpora.

The foreground and background lexicons are suited to different kinds
of WSD strategies.  For the background lexicon, statistical methods
for coarse-grained disambiguation are appropriate.  For the foreground
lexicon, WSD will occur as a by-product of finding a coherent semantic
interpretation of an input sentence, in which all arguments are of the
appropriate type.  Once the foreground/background distinction is
developed, there is a good match between what is possible, given the
state of the art in WSD and acceptable levels of human input, and what
is required, for high-quality IE.

The two-tier approach has been adopted by a
number of IE systems.  The POETIC \cite{poetic:96} and Sussex MUC-5
\cite{poetic-muc} systems used a hand-crafted foreground lexicon and
the Alvey Tools lexicon \cite{Alvey-lex:89} as a background lexicon
for syntactic information. \cite{Cahill:94} discusses the relation
between the respective roles of the two lexicons.  The Sheffield MUC-6
system \cite{Sheffield-muc6} used the Brill tagger as its background
lexicon for syntactic information.  The need for an IE system to have,
on the one hand, well-articulated meaning representations for key
terms, and on the other, some information about all or nearly all
words, makes it very likely that two-tier strategies will be adopted
even where they are not explicitly defended.

Some terminology:  I shall use `lexicographer' to refer
to the people who provide information about words, or about how words
and classes of words relate to the categories in an ontology.  At
times it might seem that `knowledge engineer' or similar is a better
description, but there is no clear point at which lexicography turns
into knowledge engineering, so I shall use the one term
throughout. Likewise, my `foreground lexicons' might equate to
Gaizauskas and Wilks's `concepticons'\footnote{See their contribution to
the SCIE Summer School, ``Concepticons {\em vs.}\ lexicons''.}
or even
knowledge representation schemes, but I shall keep to `lexicons'.
Small capitals are used to refer to semantic classes.

\section{Characteristics of IE}

For NLP tasks such as Machine Translation, Information Retrieval\footnote{The Information
Retrieval task is to return those texts, in a database of texts, which
are the most relevant to a user's query.  In contrast, IE extracts
facts from texts.} and
grammar checking, both
input and output are defined in terms of linguistic objects, so world
knowledge is in a sense optional: it is merely a means to an end.  If
statistical methods are a better means to the end, so much the better;
general knowledge can be dispensed with.  Thus world knowledge may
be useful for a task such 
as prepositional phrase attachment or
anaphor resolution, but if statistical methods perform better, then
world knowledge can be dispensed with.

The situation in relation to IE (and also for many language generation
applications) is different.  Non-linguistic objects, in the form of
templates and database fields, are part of the task definition.  If
lexical information is not tied to those objects, the task cannot be
accomplished at all.  A central problem for most knowledge-engineering
projects designed to support NLP is the lack of criteria regarding
what knowledge is relevant (see \cite{Bateman:91} for discussion).
For IE, the question arises only to a limited degree.  The templates
and database fields define what objects and relations are relevant.  

All NLP tasks are easier if only one type of text, or the language of
only one domain, is addressed, but for some tasks, including MT, IR
and grammar checking, it is theoretically feasible to produce
domain-independent systems.\footnote{Practical MT systems use
multiple, domain-specific lexicons, so if, for example, a legal text
is being translated, only legal and general-language lexicons will be
accessible: in this way, the system benefits, to some extent, from the
advantages of doing domain-specific NLP.} (There is of course
commercial pressure in this direction: a general-purpose system has a
far, far larger market.)  For IE, a completely general-purpose system
is not a coherent concept (unless various AI-complete problems are
solved and a completely general-purpose knowledge representation
scheme is available) since the database fields or templates are
domain-specific.

So: because of the way in which an IE task is
defined, firstly, an IE lexicon must include mappings to
non-linguistic objects, and, secondly, for a new domain, some
lexicography will always be required.

\section{Foreground lexicons}

While researchers in NLU have made great progress in extracting
lexical information automatically from machine-readable versions of
dictionaries (eg. \cite{WilksSlatorGuthrie,Richardson:97}) and from
text corpora (see Section~\ref{background}), these methods do not
provide the depth of knowledge about the key terms for a domain
which is required for IE.

An example: one strand of the recent MUC-6\footnote{MUC: Message
Understanding Conference.} competition concerned
`succession events', so the information to be extracted related to
individuals getting promoted, demoted, hired and fired.  A salient
term is, thus, verbal {\em sack}.  Its meaning, in the context of MUC-6, is
that the individual to whom it applies (eg., who occupies the direct
object slot, or the subject if the verb is passive) no longer has the
role he or she previously had in the organisation which either
occupies the subject slot of the active form, or whose agent occupies
that slot, or that is otherwise salient in the context, and for whom
the individual previously worked; and that the event was instigated by
the organisation rather than the individual.  Automatic
dictionary-based techniques might, if they are well done, allow us to
follow a hypernym chain from {\em sack} (verbal sense 2) to {\em
dismiss} (sense 2) to {\em remove} (sense 3),\footnote{Sense numbering
from \cite{LDOCE2}.} so supplying the fact that these three verb
senses have the same semantics in this domain.  However the step from
``same semantics'' to what that semantics is, is a large one.  
For the MUC-6 task, the semantics must specify which templates a {\sc
sack/dismiss/remove} event relates to, which slots on the template
each of the verbs' complements correspond to, the changes from the
`before' to the `after' state that the event implies, and the fact
that the employer instigated the change.
This is well beyond the potential of the kind of `shallow semantics' which
form a reasonable objective for machine-readable dictionaries or
corpus-based lexical acquisition.  

The consequence is that, for the foreseeable future, any IE project
will need to do a significant amount of lexicography.  The meanings of
the key terms in the domain, or ``foreground lexicon'' will need to be
written in a formalism which supports the reasoning the system will
need to perform and is geared to the output specifications of the IE
system.

In sum, the foreground lexicon for a domain will contain:

\begin{itemize}
\item  the key predicates for the domain;
\item how they and their arguments relate to the IE system's output
formalism; 
\item the sets of lexical items which realise the predicate; and,
\item how their complements relate to the predicate's arguments.
\end{itemize}

\subsection{WSD and the Foreground Lexicon}

The relation to word sense disambiguation has two aspects.  First,
there will probably only be one sense of {\em sack}, {\em dismiss} or
{\em remove} in the foreground lexicon.  Given a domain-specific
corpus, for many words, most or all uses of the word will be in its
foreground sense.\footnote{This is closely related to the ``One sense
per discourse'' observation, presented and quantified in
\cite{GCY:93}.}  So many words which are ambiguous in general language
are not ambiguous within the domain.  This will only be true to a
moderate degree in the MUC-6 corpus, where the input text is taken
from the Wall Street Journal so is not highly domain specific.  It
applies to a greater extent to domains such as Remote Sensing
\cite{Basili:97} or traffic information reports \cite{poetic:96}. 

Secondly, the first argument of {\sc sack/dismiss/remove} must be an
employer, and the second, an individual. These are hard constraints,
not statistical ones, as if they are violated the database entry or
template will be garbage.\footnote{Also, statistical WSD methods will 
be hard to apply, firstly because these word senses are structured
entities, secondly, there will be no training data and probably
insufficient data for unsupervised methods, and thirdly,
characterisations of each sense will not be available in a form which
is easily integrated into the algorithms.}  They will have been
added to the predicate's representation by the lexicographer.  They
will provide critical clues for disambiguation.  If the subject of
{\em sack}, {\em dismiss} or {\em remove} is correctly identified as
an employer (or agent of an employer), and its direct object, as an
individual, then just one of the three verbal senses of {\em sack},
one of the five for {\em dismiss}, and three of the five for {\em
remove} remain possible (and the two non-foreground senses for {\em
remove} which are still possible are superordinates for the foreground
sense, with the same general meaning but not applied specifically to
employment). Therefore, if we encounter {\em dismiss}, and succeed in
identifying an {\sc employer} subject (implicit or explicit) and {\sc
individual} object, we may conclude that we have the foreground sense
of {\em dismiss}.  Identifying the subject and object and their
categories is a task that must be performed in any case, in order to
ascertain how the verb's complements relate to the database or
template fields, so disambiguation has occurred without any specific
effort, as a by-product of arriving at a coherent semantic
representation for the sentence.\footnote{The three verbs are all
frequently used in their relevant sense in the (usually agent-less)
passive.  In that case, there will be fewer selection restrictions to
constrain the meaning, but on the other hand the simple fact of
passive use will implicate the foreground sense.}

If {\em dismiss} does not have an {\sc employer} subject
and {\sc individual} object, we shall not have disambiguated it
between the four non-foreground senses, but then there is no need to
do so since, whichever of those four senses applies, the verb will not
lead to information going into the templates or database.

If {\em dismiss} had another sense that had implications for the IE
task, then it would have another foreground sense.  Then three cases
are possible.  In the first, the two foreground senses are the same
concept in this domain, so we have a simple many-to-one mapping
between the dictionary senses and the domain-specific senses, and the
dictionary's sense distinction is ignored.  In the second, the two
senses relate to distinct concepts and have distinct selection
restrictions.  Once the semantic classes of the complements are
identified, the word is disambiguated, again with no explicit
disambiguation effort.  The third is the difficult case, where the two
senses relate to distinct concepts but share the same selection
restrictions.  I doubt whether this will occur often.  Where it does,
it will have been the lexicographer's task to provide sufficient
information in the concept definitions to permit disambiguation.
Since it will not occur very often, it will not be an onerous task for
the human to provide this information, given clever tools (see
Section~\ref{tools}).

Foreground lexicon disambiguation is
semantically driven: the system will know enough about the meanings of
the words and phrases for the word sense to be resolved by identifying
the only sense with a semantic fit.  This seems akin to how people
disambiguate -- not as a distinct process but as a by-product of
identifying an interpretation of the word that fits the context
\cite{Nunberg:78}.

\section{Don't be scared of lexicography}

Over the last ten years, there
have been many researcher-years spent on making information in
machine-readable dictionaries available for NLP use.  The preamble to
such work has generally included words to the effect that ``the lexicon is
huge, so if we are able to re-use existing resources, eg.\
dictionaries, we shall be making a great saving of effort''.  There
are several limitations to this argument.\footnote{See also \cite{IdeVeronis:93}.}

\begin{itemize}
\item The person-years required to make a medium-sized dictionary,
while substantial, are not necessarily forbidding.  It is likely
that more person-years have been spent extracting information
from \cite{LDOCE} than were spent in writing it.  Machine
translation laboratories frequently write dictionaries, and the {\sc Comlex}
and {\sc WordNet} projects have both done so.  Much
smaller domain-specific lexicons are not necessarily huge undertakings.

\item A purpose-built dictionary
will contain the information that is needed. Existing resources are
unlikely to.  Simple items such as word class are often available for
all words, but little else is.  Filling in gaps is likely to be
labour-intensive.

\item All dictionaries contain errors. In a computational lexicography
project, resources can be devoted to ensuring accuracy where it matters.

\end{itemize}

There will not be a huge number of concepts in the lexicon for a
particular domain, so, at, say, an average of half an hour per word
for 500 key words, where a lexicon is being built from scratch, the
process may involve two or three person-months.

A careful approach to lexicon design which exploits generalisations
has the potential to greatly speed up the
lexicography.  As pointed out above, the foreground senses of {\em
sack, dismiss} and {\em remove} all map to the same concept, though not
in identical ways.  (Someone who is sacked does not, thereafter, work
for the same employer.  This does not follow for someone who is
removed from a given post.)  A formalism is required in which all the
information common to the three verbs can be stated at a general node
for the predicate, and inherited.  Then, only the non-default
facts about each word need be stated by the lexicographer
\cite{CaEv:90} and the overhead associated with adding further words
to the lexicon, where those words behave similarly to those already
encoded, is minimal.

This inheritance-based, hierarchical approach to the lexicon is also
of benefit from a multilingual perspective.  Where lexical items of
various languages relate to the output formalism in the same way, they
can be attached to the same nodes in the hierarchy
\cite{CahillGazdar:95,Nirenburg:96,HeidKruger:96}.

\section{Background Lexicons}
\label{background}

But what of the 50,000 words which might occur in the domain corpus
and are {\bf not} in the foreground lexicon?
Syntactic information about them is required so that sentences
containing them can be parsed.  Semantic information is
required for various purposes: 

\begin{itemize}
\item general parsing problems such as prepositional
phrase attachment and disambiguation of co-ordinated constructions;
\item anaphor resolution;
\item identification of which database
fields or template slots the referent of a word might occupy  -- for
example, identifying that {\em school} is an {\sc organisation} so a
noun phrase with {\em school} as its head is a potential filler for
the {\sc employer} database field or template slot;
\item for selection restrictions on the
foreground lexicon concepts, so that in, eg.\ ``the school dismissed
\ldots'', the identification of {\em school} as {\sc organisation}
indicates the foreground sense of {\em dismiss};
\item disambiguation of the background concept word.
\end{itemize}

Note that, for all these cases, the semantic information that is
required is essentially coarse-grained classification.  We need to
know that {\em school} is (in one of its senses) {\sc organisation},
nothing more.

There are, at least for English, numerous general-language resources
which can supply some or all of the information we need for most
words.  WordNet \cite{Miller:90} provides broad word-class information
and a taxonomy of semantic classes for English, and all being well,
the EuroWordNet, German WordNet and International WordNet projects
will soon extend this to numerous other languages.  Various
machine-readable versions of monolingual and bilingual dictionaries
are more or less readily available for NLP research and development
(eg.\ from Longman, Collins, Oxford University Press, Larousse,
Bibliograf etc.), and
provide (more or less explicitly and comprehensively) morphological,
syntactic, collocational and semantic category information.  Basic
syntactic and morphological information for English, Dutch and German
is available on the CELEX CD-ROM.  Sophisticated subcategorisation
information for English verbs is available in the Alvey lexicon
\cite{Alvey-lex:89}, COMLEX-Syntax or XTAG.

Moreover there now exist numerous techniques for acquiring this sort
of information from corpora, using statistical methods, with minimal
or no lexicons required as input.  The Xerox part-of-speech tagger
\cite{Cutting:92} is one of several
language-independent taggers whose output can be used for developing
part-of-speech lexicons from scratch.
\cite{Church:89c,Hindle:90,Brown:92,Grefenstette:94,McMahon:96} present various methods,
all largely or entirely language-independent, for developing semantic
classifications.

There are also hybrid techniques which use corpora to improve, extend
or `tune' the information in lexical resources.
\cite{BriscoeCarroll:97} is one of a number of pieces of work
presenting techniques for the automatic extraction of
subcategorisation frames for verbs, given a lexicon with some
syntactic information (and a parser) as input.  (See also, eg 
\cite{HindleRooth:91,Brent:93,resnik:dissertation}, and various papers in
\cite{BogPus:93}.)  

In an IE context,
`tuning' the resource, that is, adapting it, usually by fully automatic
methods, to the language of a given corpus, is particularly salient.
An example of such work is \cite{Basili:97} who take the WordNet hierarchy;
reduce it to a far simpler, 25-way (for nouns) or 15-way (for verbs)
classification scheme; disambiguate all words which remain ambiguous
in this simplified scheme, using the domain corpus and a Bayesian
classification algorithm developed by \cite{Yarowsky:92}; and are then
able to return a `tuned' version of (very coarse-grained) WordNet, in
which senses not occurring in the domain corpus have been ejected, and
for where the remaining senses are associated with domain-specific
information which can be used for disambiguation.

This has been a brief and partial survey of a very active field.  It
serves to demonstrate that there is a large number of resources (at
least for English) and corpus-based algorithms (some language- and
lexicon-independent, others less so) for providing the semantic and
syntactic information required for the background lexicon.  The match
between what the techniques can provide, and what is required for the
background lexicon, is good.  For the background lexicon, shallow
semantics of the kind which can be automatically extracted from
lexical and corpus resources is sufficient.

\subsection{WSD in the Background Lexicon}

For fine-grained automatic WSD, with grain-size as at the
WordNet synset or LDOCE sense level,
anything over 50\% success is judged very good, and
indeed the level of agreement between two teams of human taggers was
just 57\% \cite{Ng:96}.  If IE depends on current dictionary- or
corpus-based technology for fine-grained WSD, the outlook is bleak.

So it is fortunate that  the semantic information required for
the background lexicon is just coarse-grained classification, so only
coarse-grained WSD is required.   We need to determine whether {\em
bank} refers to an organisation or not, but we are not concerned with
the distinction between the building that houses that organisation,
and the organisation itself. Here, the position is 
far rosier.  Several authors report over 90\% success.  Those results
mostly used general corpora, so the prospects for domain-specific
corpora are probably better.  Basili's (op. cit.) approach to tuning
provides a disambiguation algorithm in its own right, or could be
combined with insights from \cite{Yarowsky:95}. 

In contrast to foreground disambiguation, background disambiguation
will be surface- rather than semantics-driven, and will bear very
little relation to how people disambiguate.

\section{Tools} 
\label{tools}

The trade between lexicography and NLP flows both ways.  Lexicons are
crucial resources for NLP, and NLP can provide tools for facilitating
and improving the standard of lexicography.  

Since the advent of computers in lexicography, lexicographers have
been able to base their lexical entries on corpus evidence as never
before.  The two essential tools for a lexicographer are an editor,
for writing the entry in, and a concordancer, which gives rapid access
to all instances of a search word or pattern in a
corpus.\footnote{Here, good database technology is required since
speed is critical, the corpus will often contain several hundred
million words, and a full range of regular expressions over words,
fields associated with words (eg.\ part of speech) and sequences of
words and fields, is required.}  There are many threads to current NLP
research which could improve the lexicographic tools.  A parsed corpus
and associated search software would allow the lexicographer to search
on grammatical structures.  Semantic tagging allows him or her to use
semantic features in a search pattern.
\cite{MikheevFinch:97} presents a toolkit which
identifies those lexical, syntactic and semantic
patterns which are particularly common for the target word.
\cite{Yarowsky:95}'s WSD algorithm is well suited to
lexicographic practice, since, given a small amount of evidence about
the syntactic and collocational patterns that indicate a particular
sense for a word, it will learn further disambiguating patterns.
\cite{xkwic} and \cite{Day:97} both provide computational environments for
a lexicographer to mark up corpus instances of a word with their
characteristics (which could be word-sense).  Other techniques from NLP which have potential for forming part of an
advanced lexicographer's workbench include a number of the semantic
classification algorithms, and hybrid `lexicon-improvement' approaches
described in Section~\ref{background} above.  

A good prototype for
such an advanced workstation is described in \cite{Hector}.  Our
current work includes the integration of these techniques into a still
more advanced workstation.

As the tools for the task improve, so the manual building of
the foreground lexicon becomes a less forbidding prospect.

\section{Conclusion and open questions}

In this paper I have argued that the lexicon for an IE system should
be viewed as having two parts: a foreground lexicon, containing the
key terms for the domain, which makes the links between the words in
the text and the database fields or templates to be filled, and the
background lexicon, containing all other vocabulary.  The foreground
lexicon will be built anew, with substantial lexicographer input, for
each new application, whereas general-purposes lexical resources,
preferably tuned to the domain corpus and potentially augmented by a
range of automatic lexicon-improvement algorithms, will provide all
the information required for background lexicon entries. Project
managers need not be frightened by the prospect of doing lexicography
for each new application: the number of key terms for which lexical
entries need to be written will be quite limited, and there are
various tools to facilitate the process.

Word sense disambiguation will take quite different forms in
relation to the two parts.  For words in the background lexicons,
coarse-grained disambiguation is sufficient, and various statistical
and preference-based algorithms can be used.  For the foreground
lexicon, explicit disambiguation will rarely be an issue, as a
coherent semantic interpretation will usually only be possible with
one or zero foreground senses.

Open questions include: how large need the foreground lexicon be?  How
sharp is the distinction, and are
there intermediate cases, of word senses for which some of the
information and processing is foreground, some background?  The
discussion above suggests that background WSD would take place first,
as that would furnish the information for foreground
interpretation-building and disambiguation, but is that correct, or
how might interleaving of the processes work?  All these questions
feature as part of our programme of IE system-building.

\section*{Acknowledgements}

The paper benefited from discussions with Roger Evans, Lynne Cahill
and Robert Gaizauskas.

\bibliographystyle{fullname}

\end{document}